\documentclass[10pt,twocolumn,letterpaper]{article}

\usepackage{cvpr}
\usepackage{times}
\usepackage{epsfig}
\usepackage{graphicx}
\usepackage{amsmath}
\usepackage{amssymb}

\usepackage{multirow}

\def\ie{{\em i.e.}}
\def\eg{{\em e.g.}}
\def\etal{{\em et al. }}

\usepackage[pagebackref=true,breaklinks=true,letterpaper=true,colorlinks,bookmarks=false]{hyperref}

\cvprfinalcopy 

\def\cvprPaperID{11} 

\ifcvprfinal\fi
\begin{document}

\title{Exposing DeepFake Videos By Detecting Face Warping Artifacts}

\author{Yuezun Li, Siwei Lyu \\
Computer Science Department \\
University at Albany, State University of New York, USA
}

\maketitle

\begin{abstract}
In this work, we describe a new deep learning based method that can effectively distinguish AI-generated fake videos (referred to as {\em DeepFake} videos hereafter) from real videos.  Our method is based on the observations that current DeepFake algorithm can only generate images of limited resolutions, which need to be further warped to match the original faces in the source video. Such transforms leave distinctive artifacts in the resulting DeepFake videos, and we show that they can be effectively captured by convolutional neural networks (CNNs). 
Compared to previous methods which use a large amount of real and DeepFake generated images to train CNN classifier, our method does not need DeepFake generated images as negative training examples since we target the artifacts in affine face warping as the distinctive feature to distinguish real and fake images. The advantages  of our method are two-fold: (1) Such artifacts can be simulated directly using simple image processing operations on a image to make it as negative example. Since training a DeepFake model to generate negative examples is time-consuming and resource-demanding, our method saves a plenty of time and resources in training data collection; (2) Since such artifacts are general existed in DeepFake videos from different sources, our method is more robust compared to others.
Our method is evaluated on two sets of DeepFake video datasets for its effectiveness in practice.
\end{abstract}

\section{Introduction}
\label{sec:intro}

The increasing sophistication of mobile camera technology and the ever-growing reach of social media and media sharing portals have made the creation and propagation of digital videos more convenient than ever before. Until recently, the number of fake videos and their degrees of realism have been limited by the lack of sophisticated editing tools, the high demand on domain expertise, and the complex and time-consuming process involved. However, the time of fabrication and manipulation of videos has decreased significantly in recent years, thanks to the accessibility to large-volume training data and high-throughput computing power, but more to the growth of machine learning and computer vision techniques that eliminate the need for manual editing steps. 

In particular, a new vein of AI-based fake video generation methods known as {\em DeepFake} has attracted a lot of attention recently. It takes as input a video of a specific individual ('target'), and outputs another video with the target's faces replaced with those of another individual ('source').  The backbone of DeepFake are deep neural networks trained on face images to automatically map the facial expressions of the source to the target. With proper post-processing, the resulting videos can achieve a high level of realism. 

In this paper, we describe a new deep learning based method that can effectively distinguish DeepFake videos from the real ones. Our method is based on a property of the DeepFake videos: due to limitation of computation resources and production time, the DeepFake algorithm can only synthesize face images of a fixed size, and they must undergo an affine warping to match the configuration of the source's face. This warping leaves distinct artifacts due to the resolution inconsistency between warped face area and surrounding context. As such, this artifacts can be used to detect DeepFake Videos.

\begin{figure}[t]
	\centering
	\includegraphics[width=0.9\linewidth]{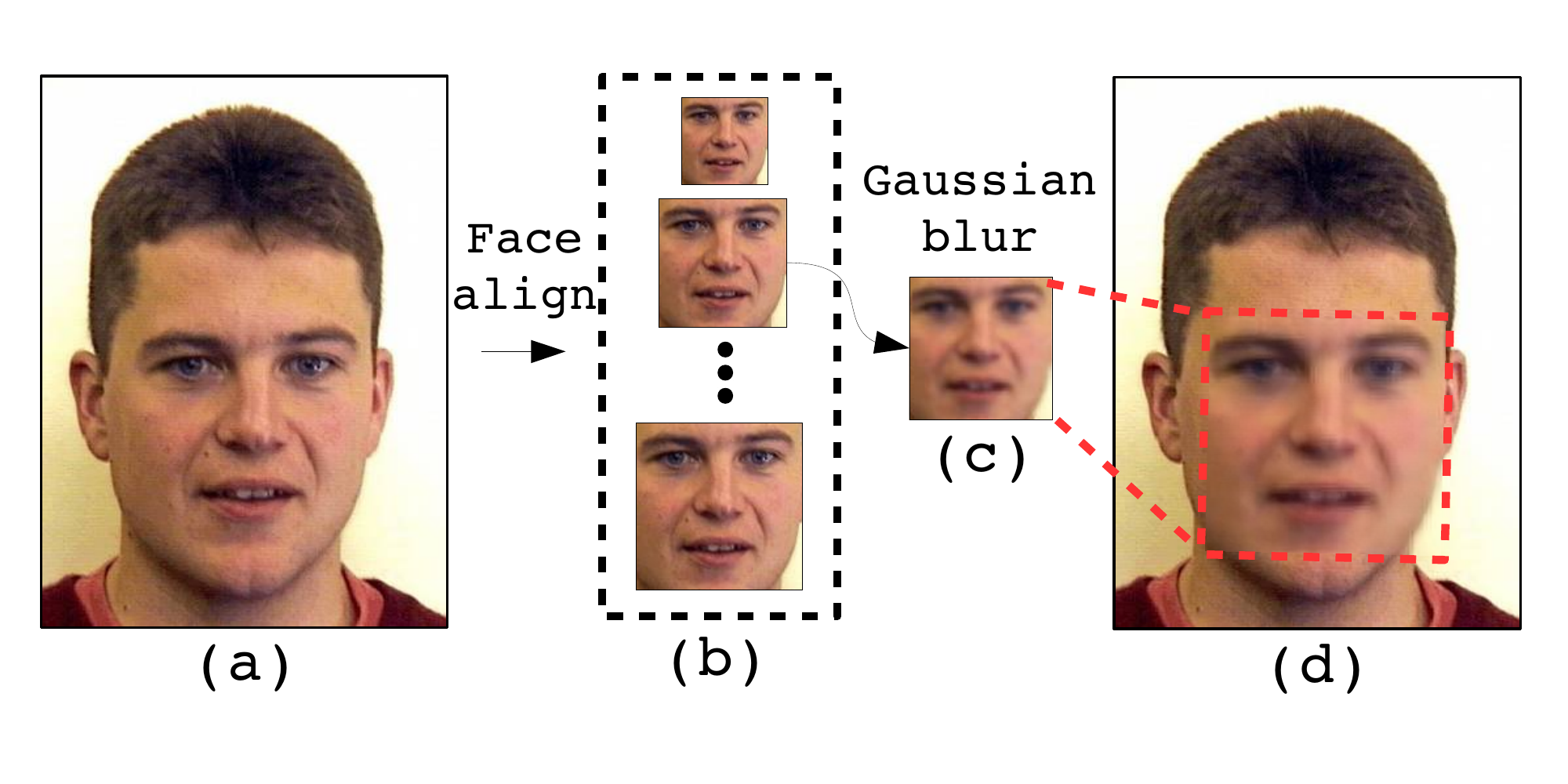}
	\vspace{-0.4cm}
	\caption{\small \em Overview of negative data generation. (a) is the original image. (b) are aligned faces with different scales. We randomly pick a scale of face in (b) and apply Gaussian blur as (c), which is then affine warped back to (d). }
	\vspace{-0.5cm}
	\label{fig:data_proc}
\end{figure}

Our method detects such artifacts by comparing the generated face areas and their surrounding regions with a dedicated Convolutional Neural Network (CNN) model.  To train the CNN model, 
we simplify the process by simulating the resolution inconsistency in affine face warpings directly. Specifically, we first detect faces and then extract landmarks to compute the transform matrices to align the faces to a standard configuration. We apply Gaussian blurring to the aligned face, which is then affine warped back to original image using the inverse of the estimated transformation matrix. In order to simulate more different resolution cases of affine warped face, we align faces into multiple scales to increase the data diversity (see Figure \ref{fig:data_proc}).
Compared to training a DeepFake model to generate fake images as negative data in \cite{afchar2018mesonet,guera2018deepfake}, which is time-consuming and resource-demanding ($\sim72$ hours on a NVIDIA GTX GPU), our method creates negative data only using simple image processing operations which therefore saves a plenty of time and computing resources. Moreover, other methods may be over-fit to a specific distribution of DeepFake videos, our method is more robust since such artifacts are general in different sources of DeepFake videos.
{Based on our collected real face images from Internet and corresponding created negative data, we train four CNN models: VGG16 \cite{simonyan2014very}, ResNet50, ResNet101 and ResNet152 \cite{he2016deep}. We demonstrate the effectiveness of our method on a DeepFake dataset from \cite{li2018ictu} and test several fake videos on {\tt YouTube}.}

\begin{figure*}[t]
	\centering
	\includegraphics[width=0.9\linewidth]{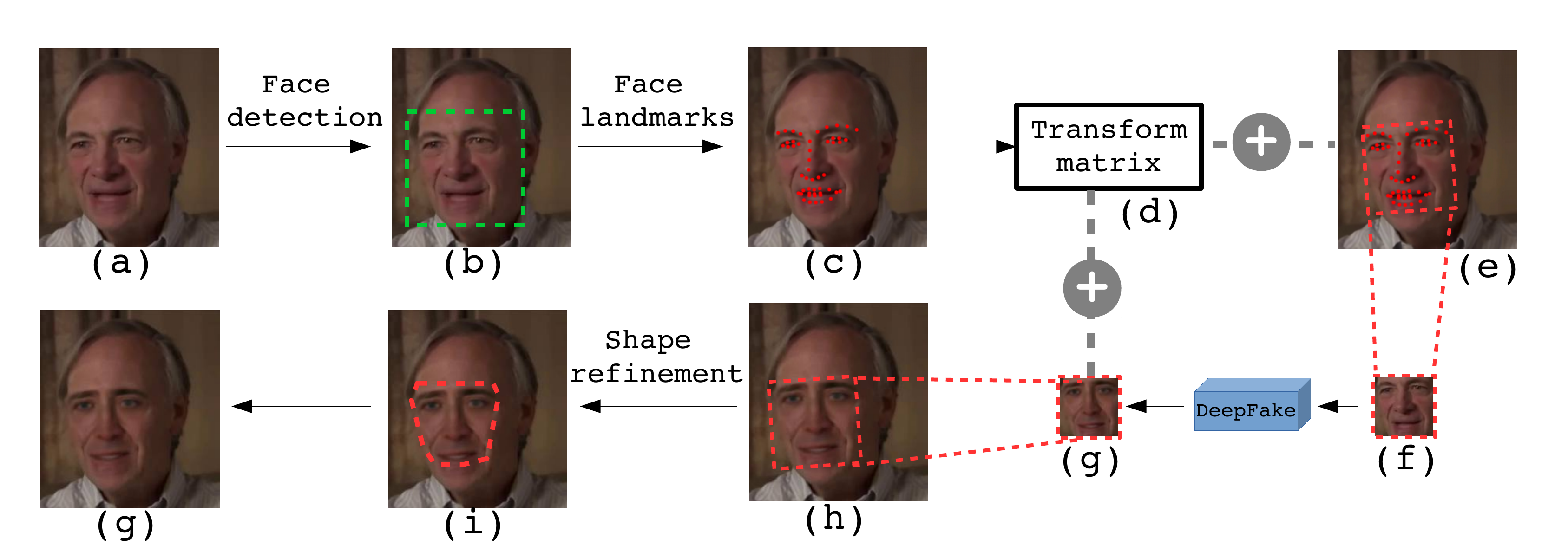}
	\caption{\small \em {Overview of the DeepFake production pipeline. (a) An image of the source. (b) Green box is the detected face area. (c) Red points are face landmarks. (d) Transform matrix is computed to warp face area in (e) to the normalized region (f). (g) Synthesized face image from the neural network. (h) Synthesized face warped back using the same transform matrix. (i) Post-processing including boundary smoothing  applied to the composite image. (g) The final synthesized image.}}
	~\vspace{-2em}
	\label{fig:overview}
\end{figure*}

\section{Related works}
\label{sec:format}

\smallskip
\noindent{\bf AI-based Video Synthesis Algorithms}
The new generation of AI-based video synthesis algorithms are based on the recent developments in new deep learning models, especially the generative adversarial networks (GANs) \cite{goodfellow2014generative}. A GAN model consists of two deep neural networks trained in tandem. The generator network aims to produce images that cannot be distinguished from the training real images, while the discriminator network aims to tell them apart. When training completes, the generator is used to synthesize images with realistic appearance.

The GAN model inspired many subsequent works for image synthesis, such as \cite{denton2015deep,radford2015unsupervised,arjovsky2017wasserstein,isola2017image,taigman2016unsupervised,shrivastava2017learning,liu2017unsupervised,CycleGAN2017,bansal2018recycle,choi2018stargan}. Liu \etal \cite{liu2017unsupervised} proposed an unsupervised image to image translation framework based on coupled GANs, which aims to learn the joint representation of images in different domains. This algorithm is the basis for the DeepFake algorithm. 

The creation of a DeepFake video starts with an input video of a specific individual ('target'), and generates another video with the target's faces replaced with that of another individual ('source'), based on a GAN model trained to translate between the faces of the target and the source, see Figure \ref{fig:overview}. More recently, Zhu \etal \cite{CycleGAN2017} proposed cycle-consistent loss to push the performance of GAN, namely Cycle-GAN. Bansal \etal  \cite{bansal2018recycle} stepped further and proposed Recycle-GAN, which incorporated temporal information and spatial cues with conditional generative adversarial networks. StarGAN \cite{choi2018stargan} learned the mapping across multiple domains only using a single generator and discriminator.

\smallskip
\noindent{\bf Resampling Detection}. The artifacts introduced by the DeepFake production pipeline is in essence due to affine transforms to the synthesized face. In the literature of digital media forensics, detecting transforms or the underlying resampling algorithm has been extensively studied, \eg, \cite{popescu2005exposing,prasad2006resampling,mahdian2008blind,kirchner2008fast,kirchner2008hiding,kirchner2009resampling,dalgaard2010role,nguyen2012robust,qian2012image,hou2014image,bunk2017detection}. However, the performance of these methods are affected by the post-processing steps, such as image/video compression, which are not subject to simple modeling. Besides, these methods usually aim to estimate the exact resampling operation from whole images, but for our purpose, a simpler solution can be obtained by just comparing regions of potentially synthesized faces and the rest of the image -- {the latter are expected to be free of such artifacts while the existence of such artifacts in the former is a telltale cue for the video being a DeepFake.}

\smallskip
\noindent{\bf GAN Generated Image/Video Detection}. Traditional forgery can be detected using methods such as \cite{zhou2017two,cozzolino2018noiseprint}. Zhou \etal \cite{zhou2017two} proposed two-stream CNN for face tampering detection. NoisePrint \cite{cozzolino2018noiseprint} employed CNN model to trace device fingerprints for forgery detection. Recently, detecting GAN generated images or videos has also made progress. Li \etal \cite{li2018ictu} observed that DeepFake faces lack realistic eye blinking, as training images obtained over the Internet usually do not include photographs with the subject's eyes closed. The lack of eye blinking is detected with a CNN/RNN model to expose DeepFake videos. However, this detection can be circumvented by purposely incorporating images with closed eyes in training. Yang \etal \cite{yang2018exposing} utilized the inconsistency in head pose to detect fake videos.
The work \cite{li2018detection} exploited the color disparity between GAN generated images and real images in non-RGB color spaces to classify them. The work \cite{mccloskey2018detecting} also analyzed the color difference between GAN images and real images. However, it is not clear if this method is extensible to inspecting local regions as in the case of DeepFake. Afchar \etal \cite{afchar2018mesonet} trained a convolutional neural networks namely MesoNet to directly classify real faces and fake faces generated by DeepFake and Face2face \cite{Thies_2016_CVPR}. The work \cite{guera2018deepfake} extended \cite{afchar2018mesonet} to temporal domain by incorporating RNN  on CNN. While it shows promising performance, this holistic approach has its drawback. In particular, it requires both real and fake images as training data, and generating the fake images using the AI-based synthesis algorithms is less efficient than the simple mechanism for training data generation in our method.

\section{Methods}
\label{sec:pagestyle}

We detect synthesized videos by exploiting the {face warping} artifacts resulted from the DeepFake production pipeline. For efficient running time, the current DeepFake algorithms create synthesized face images of fixed sizes. These faces are then undergone an affine transform (\ie, scaling, rotation and shearing) to match the poses of the target faces that they will replace ({see Figure \ref{fig:overview} (g) -- (h)}).  As such, the facial region {and surrounding regions in the original image/video frame} will present artifacts{, the resolution inconsistency} due to such transforms after the subsequent compression step to generate the final image or video frames. 
Therefore, we propose to use a Convolutional Neural Network (CNN) model to detect the presence of such artifacts from the detected face regions and its surrounding areas.  

The training of the CNN model is based on face images collected from the Internet. Specifically, we collect $24,442$ JPEG face images as positive examples. The negative examples can be generated by applying DeepFake algorithms as in \cite{afchar2018mesonet}, but it requires us to train and run the DeepFake algorithms, which is time-consuming and resource-demanding. On the other hand, as the purpose here is to detect the artifacts introduced by the {affine face warping} steps in DeepFake production pipeline, we simplify the negative example generation procedure by simulating the {affine face warping} step (Figure \ref{fig:overview}) directly.

Specifically, as shown in Figure \ref{fig:data_proc}, we take the following steps to generate negative examples to train the CNN model. 
{
\begin{enumerate}
    \item We detect faces in the original images and extract the face region using software package {\tt dlib} \cite{dlib09};
    \item We align faces into multiple scales and randomly pick one scale, which is then smoothed by a Gaussian blur with kernel size $(5 \times 5)$. This process aims to create more resolution cases in affine warped faces, which can better simulate different kinds of resolution inconsistency introduced in affine face warping.
    \item The smoothed face undergoes an affine warp back to the same sizes of original faces to simulate the artifacts in the DeeFake production pipeline.
\end{enumerate} 
}

\begin{figure}[t]
\centering
\includegraphics[width=0.9\linewidth]{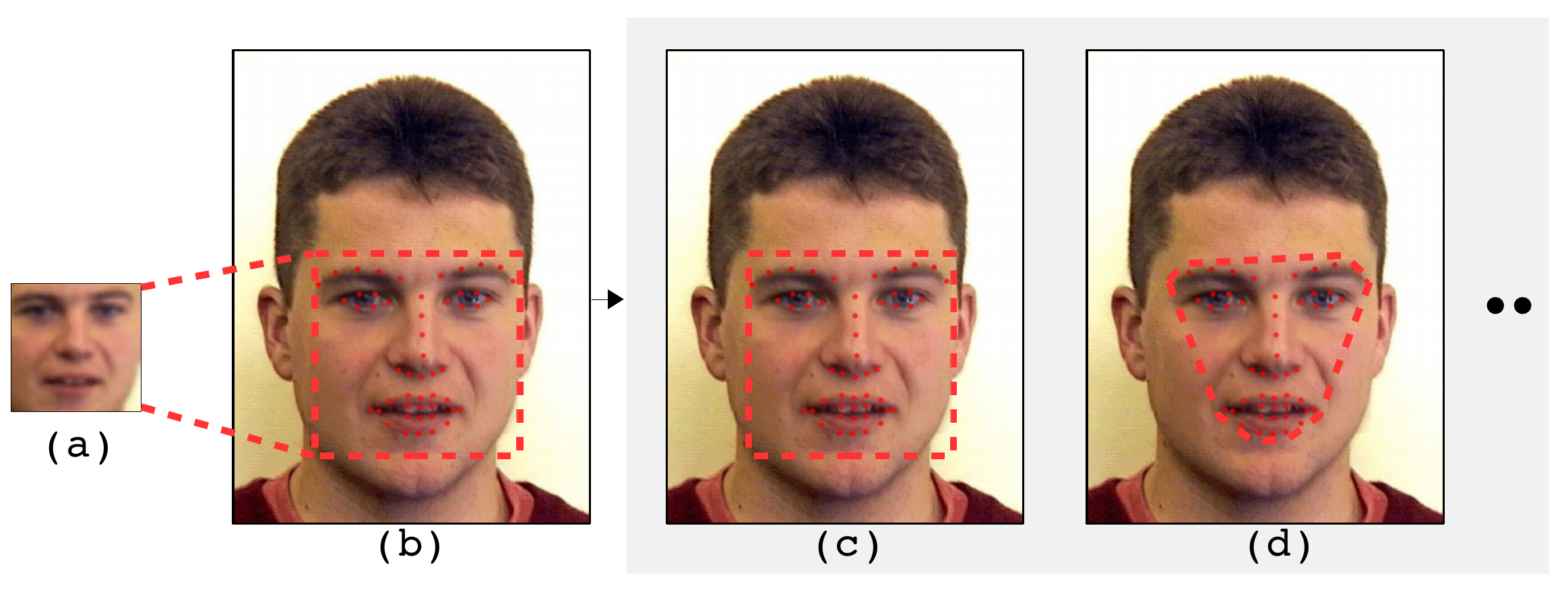}
\caption{\small \em Illustration of face shape augmentation of negative examples. (a) is the aligned and blurred face, which then undergoes an affine warped back to (b). (c, d) are post-processing for refining the shape of face area. (c) denotes the whole warped face is retained and (d) denotes only face area inside the polygon is retained.  }
\label{fig:negative_aug}
\end{figure}

To further enlarge the training diversity, we change the color information: brightness, contrast, distortion and sharpness for all training examples. {In particular, we change the shape of affine warped face area to simulate different post-processing procedure in DeepFake pipeline. As shown in Figure \ref{fig:negative_aug}, the shape of affine warped face area can be further processed based on face landmarks. Figure \ref{fig:negative_aug}(d) denotes a convex polygon shape is created based on the face landmarks of eye browns and the bottom of mouth.  }

From positive and negative examples, we crop regions of interest (RoI) as the input of our networks. As our aim is to expose the {artifacts between fake face area and surrounding area}, the RoIs are chosen as the rectangle areas that contains both the face and surrounding areas. Specifically, we determine the RoIs using face landmarks, as $[y_0 - \hat{y}_0, x_0 - \hat{x}_0, y_1 + \hat{y}_1, x_1 + \hat{x}_1]$, where $y_0, x_0, y_1, x_1$ denotes the minimum bounding box $b$ which can cover all face landmarks {excluding the outline of the cheek}. The variables $\hat{y}_0, \hat{x}_0, \hat{y}_1, \hat{x}_1$ are random value between $[0, \frac{h}{5}]$ and $[0, \frac{w}{8}]$, where $h,w$ are height and width of $b$ respectively. The RoIs are resized to $224 \times 224$ to feed to the CNN models for training.

{We train four CNN models --- VGG16 \cite{simonyan2014very}, ResNet50, ResNet101 and ResNet152 \cite{he2016deep} using our training data.
For inference, we crop the RoI of each training example by $10$ times. Then we average predictions of all RoIs as the final fake probability.}

\begin{figure}[t]
	\centering
	\includegraphics[width=0.7\linewidth]{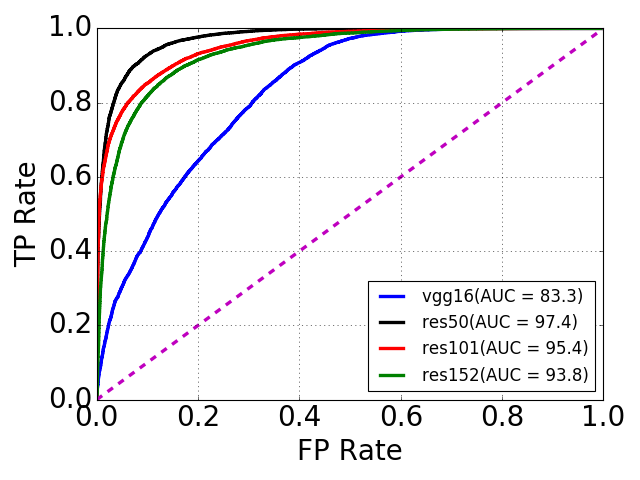}
	\caption{\small \em Performance of each CNN model on all frames of UADFV \cite{yang2018exposing}.}
	\label{fig:image_auc}
\end{figure}

\begin{figure}[t]
	\centering
	\includegraphics[width=0.7\linewidth]{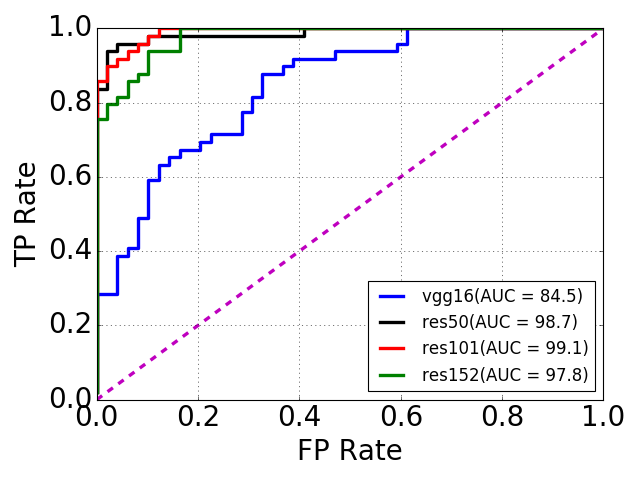}
	\caption{\small \em Performance of each CNN model on each video of UADFV \cite{yang2018exposing}.}
	\label{fig:video_auc}
\end{figure}

\section{Experiments}
\label{sec:typestyle}

We prepare our training data using the following strategy: instead of generating all negative examples in advance before training process, we employ a dynamic way to generate negative examples along with training process. For each training batch, we randomly select half positive examples and convert them into negative examples following the pipeline in Figure \ref{fig:data_proc}, which therefore makes the training data more diversified.
We set batch size as $64$, learning rate starting from $0.001$ and decay $0.95$ after each $1000$ steps. We use SGD optimization method and the training process will be terminated until it reaches the maximum epoch. 
For VGG16, we directly train it using our data and terminate it at epoch $100$. For ResNet50, ResNet101 and ResNet 152 models, we first load the ImageNet pretrained models and fine tune them using our data. The training process will be terminated at epoch $20$. Then the models are fine-tuned using hard mining strategy. In our training, hard examples include positive examples with the predicted fake probability greater than $0.5$, and negative examples with the predicted fake probability less than $0.5$. We employ the same training procedure with learning rate from $0.0001$. This stage is terminated after $20$ epochs.

\subsection{Evaluations on UADFV}
We validate our method on DeepFake video dataset {\tt UADFV} from \cite{yang2018exposing}. This dataset contains $98$ videos ($32752$ frames in total), which having $49$ real videos and $49$ fake videos respectively. Each video has one subject and lasts approximate $11$ seconds. We evaluate the four models on this dataset using Area Under Curve (AUC) metric on two settings: {\em image based evaluation} and {\em video based evaluation}.

For image based evaluation, we process and send frames of all videos into our four networks respectively. Figure \ref{fig:image_auc} illustrates the performance of each network on all frames. As these results show, the VGG16, ResNet50, ResNet101 and ResNet152 models achieve AUC performance $83.3\%, 97.4\%, 95.4\%, 93.8\%$, respectively. ResNet networks have about $10\%$ better performance compared to VGG16, due to the residual connections, which make the learning process more effective. Yet, ResNet50 has the best performance among the other ResNet networks, which shows that as the depth of network increases, the classification-relevant information diminishes.    
For video based evaluation, we take each video as the unit of analysis. {Due to the illumination changes, head motions and face occlusions in video, it is challenging to correctly predict the label of every frame. As such, we empirically assume a video is DeepFake-generated if a certain number of frames in this video are detected as fake. Thus we feed all frames of the video to the CNN based model and then return average the top third of the output score as the overall output of the video.} Figure \ref{fig:video_auc} shows the video-level performance of each type of CNN model. VGG16, ResNet50, ResNet101 and ResNet152 can achieve AUC performance $84.5\%, 98.7\%, 99.1\%, 97.8\%$ respectively. In this video based evaluation metric, ResNet network still performs $\sim 15\%$ better than VGG16. Yet, each ResNet model has similar performance, as in the case of image-level classification.  

\subsection{Evaluations on DeepfakeTIMIT}
\begin{figure}[t]
	\centering
	\includegraphics[width=0.7\linewidth]{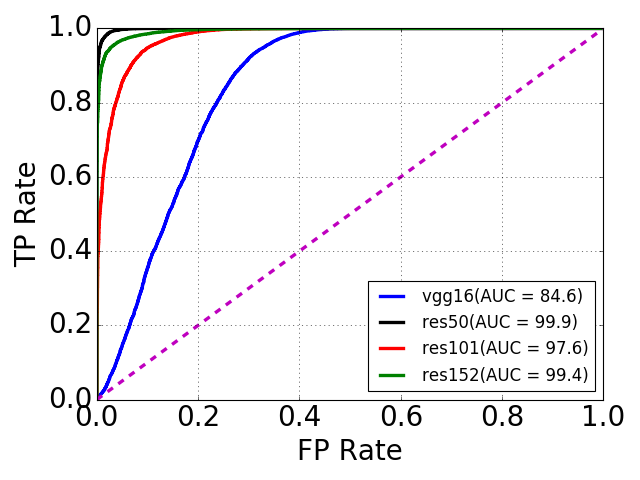}
	\vspace{-0.2cm}
	\caption{\small \em Performance of each CNN model on all frames in LQ set of DeepFakeTIMIT \cite{korshunov2018deepfakes}.}
	\label{fig:image_auc_1}
\end{figure}

\begin{figure}[t]
    \vspace{-0.5cm}
	\centering
	\includegraphics[width=0.7\linewidth]{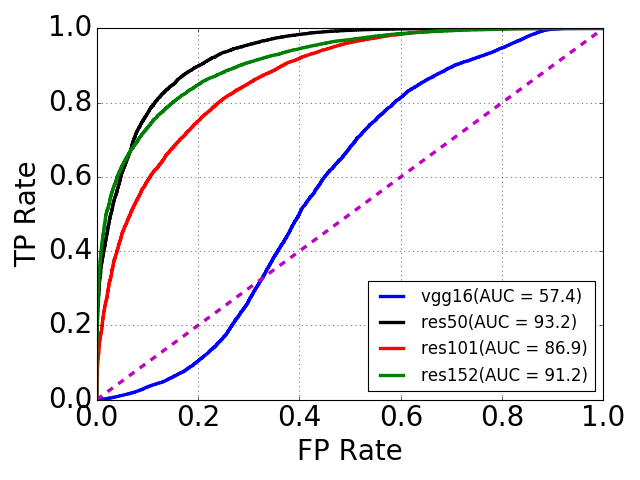}
	\vspace{-0.2cm}
	\caption{\small \em Performance of each CNN model on all frames in HQ set of DeepFakeTIMIT \cite{korshunov2018deepfakes}.}
	\label{fig:image_auc_2}
\end{figure}

In addition, we also validate our method on another DeepFake video dataset {\tt DeepfakeTIMIT} \cite{korshunov2018deepfakes}. This dataset contains two set of fake videos which are made using a lower quality (LQ) with 64 x 64 input/output size model and higher quality (HQ) with 128 x 128 size model, respectively. Each fake video set has $32$ subjects, where each subject has $10$ videos with faces swapped. Each video is $512\times384$ and lasts $\sim 4$ seconds. The original videos of corresponding $32$ subjects are from {\tt VidTIMIT} dataset \cite{sanderson2009multi}. We select subset of each subject from original dataset {\tt VidTIMIT} and all fake videos from {\tt DeepfakeTIMIT} for validation ($10537$ original images and $34023$ fake images for each quality set). We evaluate our four models on each frame of all videos based on AUC metric, where the performance of VGG16, ResNet50, ResNet101 and ResNet152 models on LQ and HQ video sets are $84.6\%, 99.9\%, 97.6\%, 99.4\%$ and $57.4\%, 93.2\%, 86.9\%, 91.2\%$ respectively, see Figure \ref{fig:image_auc_1} and Figure \ref{fig:image_auc_2}.

\begin{figure}[t]
	\centering
	\includegraphics[width=0.8\linewidth]{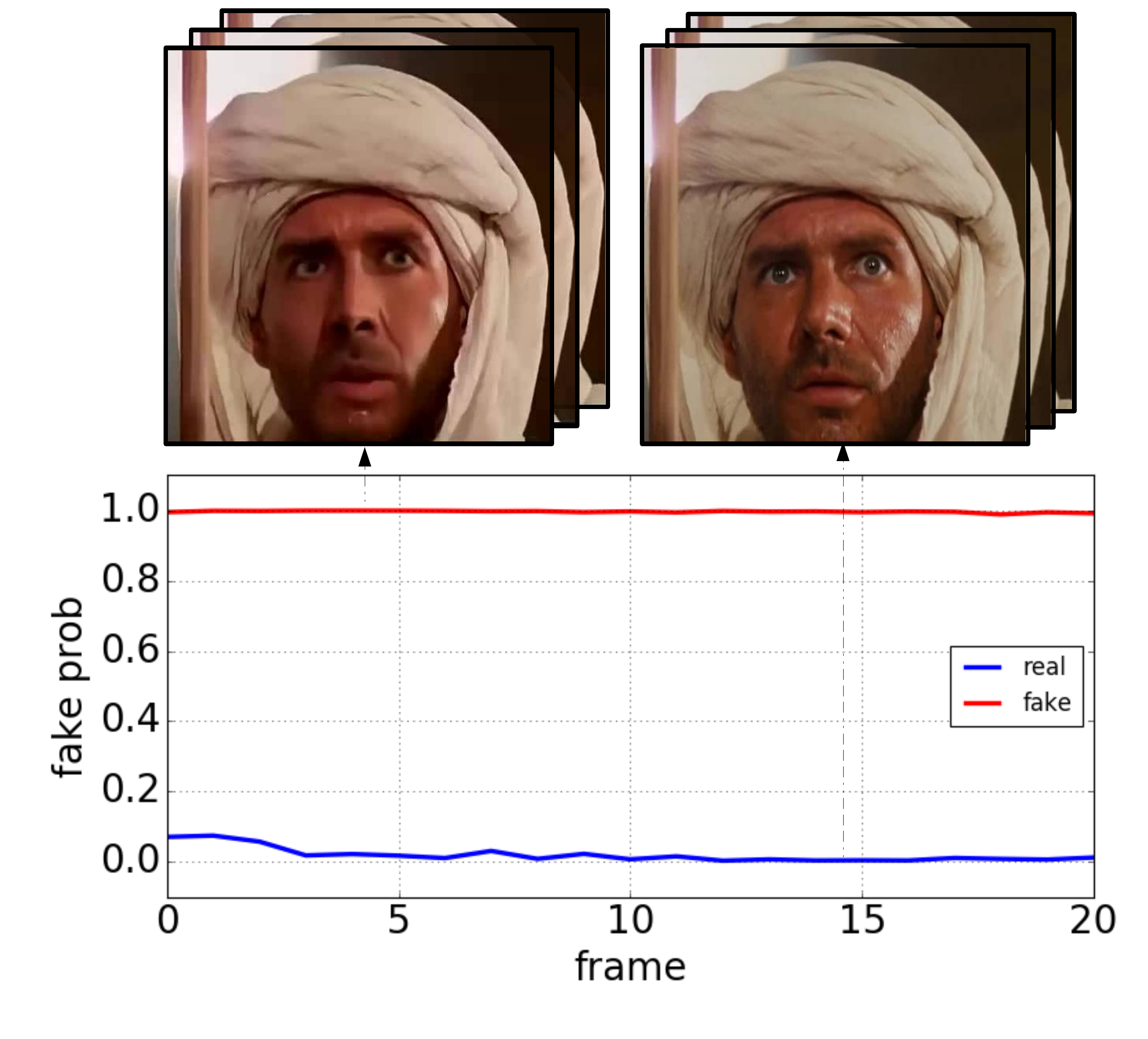}
	\vspace{-0.2cm}
	\caption{\small \em Example of our method on a DeepFake generated video clip from {\tt YouTube} ({\bf left)} and original video clip ({\bf right)}.}
	\label{fig:YouTube}
\end{figure}

We have also tested our algorithm on several DeepFake videos that were generated and uploaded to {\tt YouTube} by anonymous users. In Figure \ref{fig:YouTube}, we show the detection results as the output score from the ResNet50 based CNN model for one particular example\footnote{\url{https://www.youtube.com/watch?v=BU9YAHigNx8} }, where an output of $0$ corresponds to a frame free of the warping artifacts. As these results show, the CNN model is effective in detecting the existence of such artifacts, which can be used to determine if these videos are synthesized using the DeepFake algorithm. 

\subsection{Comparing with State-of-the-arts}

\begin{table}
\caption{AUC performance of our method and other state-of-the-art methods on UADFV and DeepfakeTIMIT datasets.}
  \begin{tabular}{l|c|cc}
    \hline
    \multirow{2}{*}{\bf Methods} & \multirow{2}{*}{\bf UADFV} & \multicolumn{2}{c}{{\bf DeepfakeTIMIT}} \\
                                 &               & LQ & HQ \\
    \hline
    Two-stream NN \cite{zhou2017two} & 85.1 & 83.5 & 73.5\\
    \hline
    Meso-4 \cite{afchar2018mesonet} & 84.3 & 87.8 & 68.4\\
    MesoInception-4 & 82.1& 80.4 & 62.7\\
    \hline
    HeadPose \cite{yang2018exposing} & 89.0 & - & -\\
    \hline
    Ours-VGG16 & 84.5 & 84.6 & 57.4\\
    Ours-ResNet50 & \bf 97.4 & \bf 99.9 & \bf 93.2\\
    Ours-ResNet101 & 95.4 & 97.6 & 86.9\\
    Ours-ResNet152  & 93.8 & 99.4 & 91.2\\
    \hline
  \end{tabular}
  \label{table:compare}
  \vspace{-0.5cm}
\end{table}

We compare the AUC performance of our method with other state-of-the-art methods: the face tampering detection method {\em Two-stream NN} \cite{zhou2017two}, and two DeepFake detection methods {\em MesoNet} \cite{afchar2018mesonet} and {\em HeadPose} \cite{yang2018exposing} on the {\tt UADFV} dataset and {\tt DeepfakeTIMIT} dataset. 
For {\em MesoNet}, we test the proposed two architectures: Meso-4 and MesoInception-4.
Table \ref{table:compare} shows the performance of all the methods. As the results show, our ResNet models outperform all other methods. Specifically, ResNet50 achieves best performance, which outperforms {\em Two-stream NN} by $\sim 16\%$ on both datasets that thereby demonstrates the efficacy of our method on DeepFake video detection. Our method also outperforms Meso-4 and MesoInception-4 by $\sim 17\%$ and $\sim 21\%$ on both datasets. Specifically, our method has a notable advance in HQ set of {\tt DeepfakeTIMIT}. Since {\em MesoNet} is trained using self-collected DeepFake generated videos, it may over-fit to a specific distribution of DeepFake videos in training. In contrast, our method focuses on more intuitive aspect in DeepFake video generation: resolution inconsistency in face warping, which is thereby more robust to DeepFake videos of different sources. {\em HeadPose} utilizes head pose inconsistency to distinguish real and fake videos. However, such physiological signal may not be notable in frontal faces, such that our method outperforms it by $\sim 8\%$ on {\tt UADFV}.

\section{Conclusion}
In this work, we describe a new deep learning based method that can effectively distinguish AI-generated fake videos (DeepFake Videos) from real videos.  Our method is based on the observations that current DeepFake algorithm can only generate images of limited resolutions, which are then needed to be further transformed to match the faces to be replaced in the source video. Such transforms leave certain distinctive artifacts in the resulting DeepFake Videos, which can be effectively captured by a dedicated deep neural network model.  We evaluate our method on several different sets of available DeepFake Videos which demonstrate its effectiveness in practice.  

As the technology behind DeepFake keeps evolving, we will continuing improve the detection method.  First, we would like to evaluate and improve the robustness of our detection method with regards to multiple video compression. Second, we currently using predesigned network structure for this task (\eg, resnet or VGG), but for more efficient detection, we would like to explore dedicated network structure for the detection of DeepFake videos.

\section*{Acknowledgment}

This research was developed with funding from Microsoft, a Google Faculty Research Award, and the Defense Advanced Research Projects Agency (DARPA FA8750-16-C-0166). The views, opinions, and findings expressed are those of the authors and should not be interpreted as representing the official views or policies of the Department of Defense or the U.S. Government. 

{\small
\bibliographystyle{ieee_fullname}
\bibliography{egbib}
}

\end{document}